\documentclass[10pt,twocolumn,letterpaper]{article}

\usepackage{iccv}
\usepackage{times}
\usepackage{epsfig}
\usepackage{graphicx}
\usepackage{amsmath}
\usepackage{amssymb}

\usepackage{booktabs, nicefrac}
\usepackage{multirow,xcolor}
\usepackage{algorithm}
\usepackage{algorithmic}
\usepackage{bm}
\usepackage{caption}

\usepackage[pagebackref=true,breaklinks=true,letterpaper=true,colorlinks,bookmarks=false]{hyperref}

\usepackage[accsupp]{axessibility}

\iccvfinalcopy %

\ificcvfinal\fi

\newcommand\gray[1]{\textcolor{gray}{#1}}

\begin{document}

\title{FrozenRecon: Pose-free 3D Scene Reconstruction with Frozen Depth Models\thanks{
First two authors contributed equally. GX is now with Zhejiang University and his contribution was made when visiting Zhejiang University.
}
}

\author{
Guangkai Xu$ ^{1*}$,
~~~
Wei Yin$ ^{2*}$,
~~~
Hao Chen$ ^{3}$,
~~~
Chunhua Shen$ ^3$,
~~~
Kai Cheng$ ^1$, 
~~~
Feng Zhao$ ^1$
\\[0.2cm]
\normalsize 
$^1$ University of Science  and  Technology of China
~
$ ^2$ DJI Technology
~
$ ^3$ Zhejiang University 
}

\makeatletter
\let\@oldmaketitle\@maketitle%
\renewcommand{\@maketitle}{\@oldmaketitle%
 \centering
    \includegraphics[width=\textwidth]{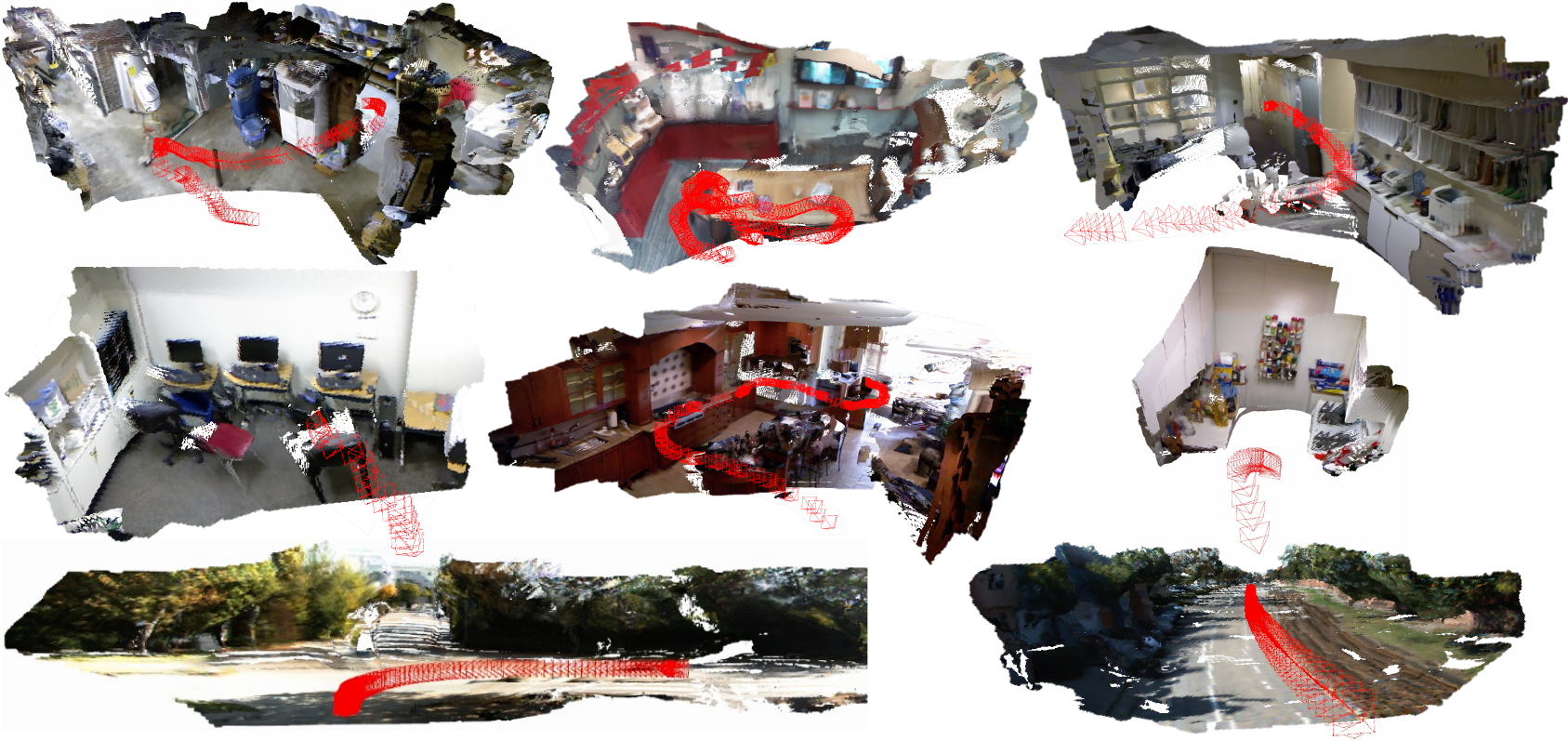}
    \vspace{-1.5 em}
    \captionof{figure}
    {
    \textbf{Robust 3D scene reconstruction of challenging and diverse scenes.} Given a monocular video, our algorithm 
    reconstructs the 3D scene without requiring 
    offline-acquired 
    camera poses. 
    Significantly, our approach only involves dozens of parameters of each frame to optimize online.
    Here, the red triangulations %stand for 
    denote 
    the 
    estimated 
    camera trajectory.
    }
    \bigskip}                   
\makeatother
\maketitle

% \ificcvfinal\thispagestyle{empty}\fi

\begin{abstract}
3D scene reconstruction is a long-standing
vision task. 
Existing approaches  can be 
categorized into geometry-based and learning-based methods. The former leverages multi-view geometry
but can face catastrophic failures due to the reliance on accurate 
pixel 
correspondence across views.
The latter was proffered to mitigate these issues by learning 2D or 3D representation directly. However, without a 
large-scale 
video or 3D training data, it can hardly generalize to diverse real-world scenarios due to the presence of tens of millions or even billions of optimization parameters in the deep network.

Recently, robust monocular depth estimation models trained with large-scale datasets 
have been proven to possess weak 3D geometry prior, but they are insufficient for reconstruction due to the unknown camera parameters, the affine-invariant property, and inter-frame inconsistency. 
Here, 
we propose a novel test-time optimization approach that can transfer the robustness of affine-invariant depth models such as LeReS to challenging diverse scenes while ensuring inter-frame consistency, with only dozens of parameters to optimize per video frame. Specifically, our approach involves freezing the pre-trained affine-invariant depth model's depth 
predictions, 
rectifying them by optimizing the unknown scale-shift values with a geometric consistency alignment module, and employing the resulting scale-consistent depth maps to robustly obtain camera poses and achieve dense scene reconstruction, even in low-texture regions. Experiments show that our method achieves state-of-the-art cross-dataset reconstruction on five zero-shot testing datasets. 
Project webpage is at: 
\href{https://aim-uofa.github.io/FrozenRecon/}{\rm https://aim-uofa.github.io/FrozenRecon/}

\end{abstract}

\vspace{-1 em}
\section{Introduction}
\label{sec:intro}

Dense 3D reconstruction is a 
fundamental vision task,  
which has a wide range of applications in autonomous driving  
\cite{%
coenen2019precise}, virtual$/$augmented reality   \cite{bruno20103d, yang2013image}, robot navigation,
medical-CAD modeling  \cite{%
sun2005bio}, \etc. 
Existing geometry-based methods and learning-based methods have achieved impressive performance. 
Despite progress, we observe 
failure cases in certain real-world scenarios, such as incomplete or noisy reconstruction of low-texture scenes, tracking failures for pose estimation, limited generalization to %new
unseen 
scenes, \textit{etc}. 
In this work, \textit{we aim to robustly and efficiently reconstruct  diverse scenes on monocular videos, with the intrinsic camera parameters and poses jointly optimized at the same time. 
}

Most 3D scene reconstruction algorithms 
can be categorized 
into learning-based %
  \cite{murez2020atlas, sun2021neuralrecon, xie2022planarrecon} and geometry-based 
methods
  \cite{zhou2017unsupervised, bian2019unsupervised, bian2021auto, im2018dpsnet, yao2018mvsnet, luo2020consistent, kopf2021rcvd, lee2022globally}. 
The first group of methods 
typically
rely 
on a 
strong
neural network to learn the scene geometry, which often requires large-scale and high-quality data to optimize
for 
millions or billions of learnable parameters. Such an expensive data requirement limits 
applications to various scenarios. Furthermore, some methods, such as SC-DepthV3  \cite{sun2022sc}, optimize the depth and poses at the same time, for which the optimization can be challenging and may 
become stuck in trivial solutions, partly due to the large number of 
optimization 
parameters.  

In contrast, geometry-based methods 
find feature correspondences  
across views 
to achieve dense 3D reconstruction. Thus, fewer$/$no training data are needed. 
The drawback is that these methods 
can easily fail 
on texture-less or low-texture scenes. Moreover, planar scenes and in-place rotations also lead to 
degeneration 
in camera pose optimization. 
Occlusion, lighting changes, and low-texture regions can make the dense matching intractable and thus lead to incomplete reconstructions. Seeking to %
mitigate these limitations, we propose to 
exploit a pre-trained, robust monocular depth model to obtain 
scene geometry priors, which can largely 
ease pose optimization and reconstruction.

Recently, foundation models  \cite{radford2021learning, alayrac2022flamingo} trained with large-scale datasets can generalize to new datasets/tasks with few samples by optimizing for a small portion of parameters (so-called adaptors). Inspired by the success, we 
use 
a  pre-trained monocular depth model of 
strong performance 
such as LeReS  \cite{yin2021learning}, 
and %
optimize 
for 
a very sparse set of parameters for quickly  
rectifying the depth maps on test videos, such that scale-consistent depth maps can be attained.

Models such as LeReS  \cite{yin2021learning}, MiDaS  \cite{ranftl2020towards}, and DPT  \cite{ranftl2021vision} 
use 
millions of training 
images 
to train a robust monocular depth model, which generalizes well to diverse scenes. 
Unfortunately, the 
predicted depth maps of those models are affine-invariant, \textit{i.e.}, up to an unknown scale and shift compared against the ground-truth metric depth. As pointed out in   \cite{yin2021learning},  the unknown shift can cause significant distortion. Naively fusing pre-frame prediction is prone to 
cause reconstruction distortion and scale misalignment. Furthermore, the unknown intrinsic camera parameters and poses are another obstacle %
for 
multi-frame reconstruction.
If we have access to 
accurate camera parameters, we can fine tune the pre-trained model to 
solve the above issues, as  in   \cite{luo2020consistent}.

Instead, we 
freeze the monocular depth model, %
and on the given videos, we  
optimize %
for 
the global scale value, the global shift value,
a local %
scale map and a local shift map to rectify each predicted affine-invariant depth map. %Camera intrinsic 
Intrinsic camera 
parameters and camera poses are also optimized at the same time.
\textit{The  
number of parameters that need to be optimized online is only around 
30 
per 
frame. 
} This is a sharp contrast compared with learning-based approaches, \eg,  SC-DepthV3, where tens of millions of parameters are involved in optimization.
The %
shift and scale
parameters are optimized 
by supervising the photometric consistency and geometric consistency between %
selected keyframes.
Due to the sparsity of 
optimization 
parameters and the robustness of the monocular depth model, 
our method works much better in terms of domain gap compared to existing deep learning methods for this task. At the same time, 
compared with %
traditional 
geometry-based 3D scene reconstruction methods, ours can be more robust to low-texture regions.

In order to validate the robustness of our system, we 
test on $5$ unseen datasets: NYU  \cite{silberman2012indoor}, ScanNet  \cite{dai2017scannet}, 7-Scenes  \cite{shotton2013scene}, TUM  \cite{sturm2012benchmark}, and KITTI  \cite{geiger2013vision}. Experiments show that our pipeline %
outperforms %
recent methods 
and achieves state-of-the-art reconstruction performance. Besides, we also perform extensive ablation studies to explore the %
usefulness 
of components of our reconstruction pipeline. Our main contributions are summarized as follows:
\begin{itemize}
\itemsep -0.05cm 
    \item We propose a novel pipeline for dense 3D scene reconstruction by
    using a frozen affine-invariant depth model, and 
    jointly optimizing %
    a sparse %
    set of 
    parameters for rectifying depth maps, camera poses, and intrinsic camera parameters.
    Our method 
    is robust on diverse unseen scenarios.

\item 
Experiments on diverse datasets show the robustness of our method and verify the usefulness of each component in our method. 
  
\end{itemize}

\begin{figure*}[t!]
\centering    %
\includegraphics[width=\linewidth]{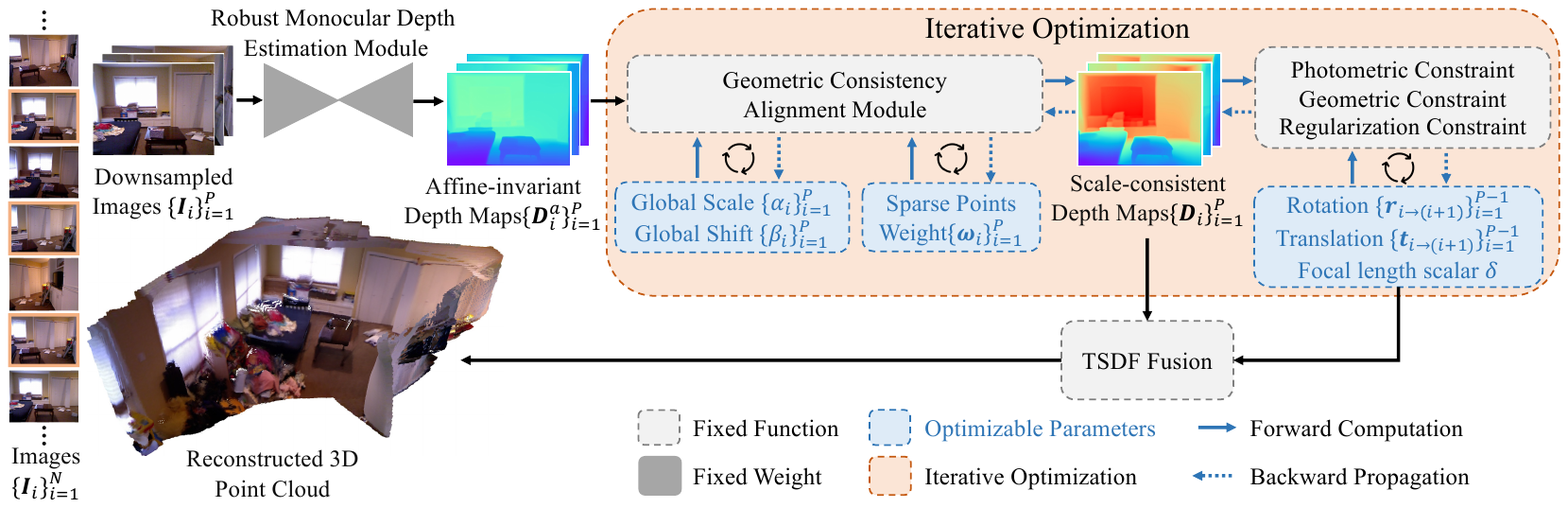}
%\vspace{-0.5 em}
\caption{\textbf{Pipeline.} Given a monocular video, we use a frozen, robust monocular depth estimation model to obtain the estimated depths of all frames. 
Then, we propose a geometric consistency alignment module, which optimizes a sparse set of parameters (\textit{i.e.}, scale, shift, and weight factors) to achieve multi-view geometric consistent depths among all frames. 
The camera's intrinsic parameters and poses are also optimized simultaneously.  
As a result, we 
achieve high-quality dense 3D reconstruction with optimized depths and camera parameters. 
} %
\label{fig: pipeline}
%\vspace{-1.5 em}
\end{figure*}

\section{Related Work}

\noindent\textbf{Learning-based %
3D Scene Reconstruction.} Learning-based methods  \cite{murez2020atlas, sun2021neuralrecon, xie2022planarrecon} 
rely on a neural network with a large number of learnable parameters to learn the geometry of %
scenes. 
Some approaches generate the 3D voxel volume from the 2D image features of the entire sequence  \cite{murez2020atlas} or local fragments  \cite{sun2021neuralrecon}, and estimate an implicit 3D representation from posed images. 
Besides, some other algorithms  \cite{zhou2017unsupervised, bian2019unsupervised, bian2021auto} attempt to predict poses and depth maps jointly. They 
% propose to
employ an ego-motion network and a depth network to predict relative camera poses and depth maps, and supervise the photo-metric consistency  \cite{zhou2017unsupervised, bian2021auto} and geometric consistency   \cite{bian2021auto}.

\noindent\textbf{Multi-view Geometry Based 3D Scene Reconstruction.} 
Multi-view geometry based methods  \cite{zhou2017unsupervised, bian2019unsupervised, bian2021auto, im2018dpsnet, yao2018mvsnet, cheng2022exploiting, huang2018deepmvs} 
achieve dense 3D reconstruction by finding feature correspondences across views. 
Traditional methods  \cite{schonberger2016structure, schonberger2016pixelwise} extract 
image 
features, establish correspondences between frames, 
and 
optimize the pose and depth iteratively, typically using bundle adjustment  \cite{triggs1999bundle}.
To achieve robust feature representation, some 
methods leverage plane sweeping to establish correspondences with assumed depth values. They usually construct a cost volume with the CNN-extracted features between images \cite{huang2018deepmvs, yao2018mvsnet, im2018dpsnet} 
, and employ a ConvNet to predict depth maps after regularization. 
Compared with those approaches, 
ours is 
less likely to suffer from pose catastrophic failures in low-texture regions and can be more robust to diverse scenes.

Besides, %
visual SLAM
systems 
  \cite{mur2015orb, mur2017orb, campos2021orb, engel2017direct, engel2014lsd, forster2014svo,
sarlin2020superglue, teed2021droid}  also
estimate the camera motion and construct the map of unknown environments. %
Methods of 
  \cite{mur2015orb, mur2017orb, campos2021orb, engel2017direct} rely on the assumption of geometric or photometric consistency 
to estimate poses and depth through bundle adjustment. Recent works  \cite{
sarlin2020superglue, teed2021droid} propose to improve the robustness and accuracy by integrating more representative clues and optimization tools from deep learning. 
These algorithms focus on accurate pose estimation, but may fail to build dense maps with geometry details.

\noindent\textbf{Robust Monocular Depth Estimation.}
To achieve robust monocular depth estimation, 
some methods   \cite{%Yin_2023_ICCV, 
ranftl2020towards, ranftl2021vision, yin2021learning, %xu2022boosting, 
yin2021virtual} 
learn the affine-invariant depth with large-scale datasets,  which is more likely to be robust to unseen scenarios but up to an unknown scale and shift. Although 
they achieve 
promising robustness, %
the estimated affine-invariant depth needs to recover scale-shift values by globally aligning  \cite{ranftl2020towards, ranftl2021vision, yin2021learning} 
with ground-truth depth. Some algorithms  \cite{luo2020consistent, kopf2021rcvd, lee2022globally} propose to employ robust depth prediction and achieve visually consistent video depth estimation. In contrast, this work focuses on leveraging robust monocular depth models and aiming at dense 3D scene reconstruction without extra offline-obtained information.

\section{Method}

\begin{figure*}[t!]
\centering    %
\includegraphics[width=\linewidth]{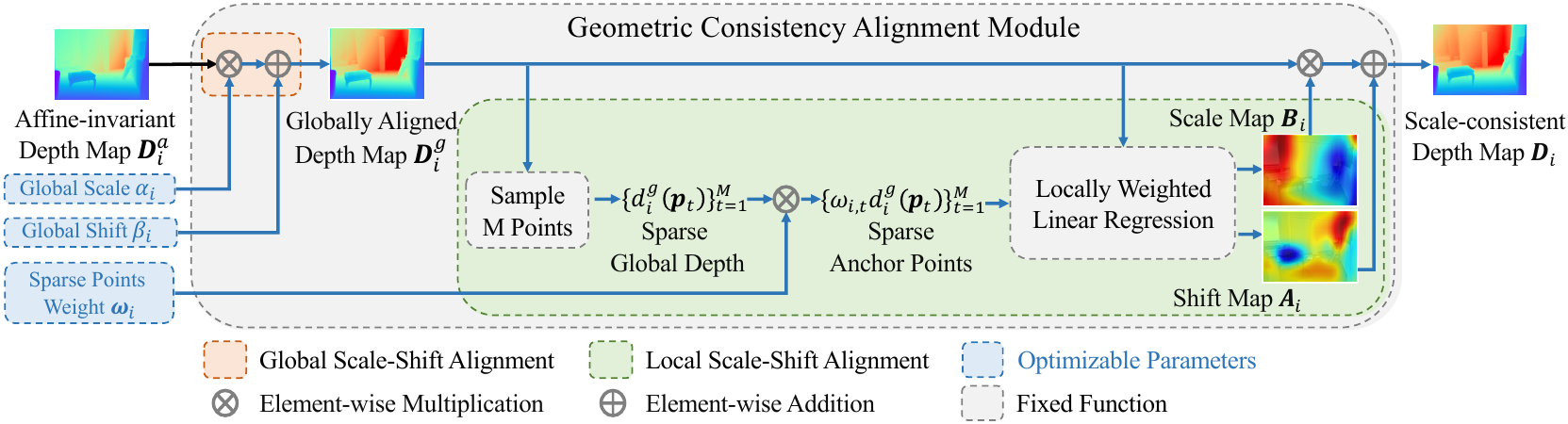}
%\vspace{-1 em}
\caption{\textbf{The geometric consistency alignment module.} We propose to recover the global scale and shift of depth map first, then 
apply 
the locally weighted linear regression method to compute the scale map and shift map, and finally achieve scale-consistent depth.
} %
\label{fig: alignment_module}
\vspace{-1 em}
\end{figure*}

\subsection{Overview}
 Aiming at 3D reconstruction from a monocular video, we propose a lightweight optimization pipeline to jointly optimize depth maps, camera poses, and intrinsic camera parameters   in Figure~\ref{fig: pipeline}. The core components of our method are the geometric consistency alignment module, optimization objectives, camera parameters initialization, and the keyframe sampling strategy.
 
 With the sampled images $\{\mathbf{I}_{i}\}_{i=1}^P$, we first use LeReS  \cite{yin2021learning, yin2022reconstruction} to obtain affine-invariant depth maps $\{\mathbf{D}^{a}_{i}\}_{i=1}^P$. To avoid disastrous point cloud duplication and distortion, we leverage a geometric consistency alignment module $\mathcal{F}(\cdot, \cdot, \cdot, \cdot)$ to retrieve unknown scale and shift of affine-invariant depth and compute scale-consistent depth $\mathbf{D}_i$:
 \begin{equation}
\label{eq: metric depth}
\begin{gathered}
    \mathbf{D}_i = \mathcal{F}(\mathbf{D}^{a}_i; 
    \; 
    \alpha_i, \beta_i, \bm{\omega}_i),
\end{gathered}
\end{equation}
where $\alpha_i$, $\beta_i$, and $\bm{\omega}_i$ are the optimization variables. 

In order to transfer the robustness of the depth model, 
three optimization objectives are supervised to ensure multi-frame consistency by warping from source frame $i$ to reference frame $j$. With this warping process, we propose to minimize the color and depth difference between warped points and reference points, \textit{i.e.}, photometric and geometric constraints. 
An additional  regularization constraint is also employed to stabilize the optimization process.

During optimization, the consistency supervision relies on two conditions: 
1) the warped reference image and source images 
must 
have overlaps; %
2) nicely overlapped frames can be automatically selected as our keyframes. %
Therefore, the appropriate initialization of camera parameters and the keyframes' sampling strategy 
are 
essential.

Finally, with the optimized intrinsic camera  parameters, camera poses, and %
scale-consistent depth maps, we can achieve accurate 3D scene reconstruction with a simple TSDF fusion  \cite{zeng20173dmatch}.

\subsection{Optimization}
\label{sec: our algorithm}

We assume a simple pinhole camera model, and use homogeneous representation for all image coordinates. By default, the scale factor in homogeneous coordinate is 1.

\noindent\textbf{Geometric Consistency Alignment Module.} %
The details of geometric consistency alignment module $\mathcal{F}(\cdot)$ are shown in Figure \ref{fig: alignment_module}. 
The predicted depths $\{\mathbf{D}^{a}_{i}\}_{i=1}^P$ are affine-invariant.
The unknown scale and shift will cause duplications and 
distortions if they are incorrectly estimated. 
To retrieve them,
we perform a two-stage alignment, \ie,
a global scale-shift alignment, and a local scale-shift alignment. In the global alignment  \cite{yin2021learning}, we optimize
for 
a global scale $\alpha_i$ and a global shift $\beta_i$ to 
obtain 
the globally aligned depth map $\mathbf{D}^{g}_{i}$ %
as follows:
\begin{equation}
\label{eq: global alignment}
\begin{gathered}
    \mathbf{D}^{g}_i = \alpha_i  \mathbf{D}^{a}_i + \beta_i.  \\
\end{gathered}
\end{equation}

In the local alignment,
we compute a scale map $\mathbf{A}_i \in \mathbb{R}^{H\times W}$ and shift map $\mathbf{B}_i \in \mathbb{R}^{H\times W}$ for each frame $i$ to 
refine the 
globally aligned depth. Instead of directly optimizing these two maps, we uniformly sample $M$ sparse global depth $\{d^{g}_i ( {\mathbf{p}_{t}} )\}^{M}_{t=1}$ from the globally aligned depth $\mathbf{D}^{g}_{i}$:
\[ 
d^{g}_i ( {\mathbf{p}_{t}} ) = f_{\rm s}(\mathbf{D}^{g}_i, \mathbf{p}_{i, t}),  \,\,\,\, {\rm for } \,\, t = 1...M.
\]
where $f_{\text{s}}(\cdot, \cdot )$ %
%samples
retrieves 
the depth value at a position; 
\eg,
$ f_{\rm s}(\mathbf{D}^{g}_i, \mathbf{p}_{i, t}) $ obtains 
$d^{g}_i ( {\mathbf{p}_{t}} )$ from $\mathbf{D}_i^g$
at the position $\mathbf{p}_{i,t}$. Then, by multiplying sparse global depth $\{d^{g}_i ( {\mathbf{p}_{t}} )\}^{M}_{t=1}$ with the weights $\{\omega_{i, t}\}_{t=1}^{M}$, we compute the sparse anchor points $\{\omega_{i, t}\cdot d^{g}_i ( {\mathbf{p}_{t}} )\}^{M}_{t=1}$ and
use the locally weighted linear regression method\cite{xu2022boosting},
\textit{i.e.}, $f_{\text{LWLR}}(\cdot, \cdot )$, to compute the local scale map $\mathbf{A}_i \in \mathbb{R}^{H \times W}$ and local shift map $\mathbf{B}_i \in \mathbb{R}^{H \times W}$. We also describe the $f_{\text{LWLR}}(\cdot, \cdot )$ module in detail in the supplementary. Such indirect optimization %is to 
reduces the parameters to ease optimization. The local alignment is as follows.
\begin{equation}
\label{eq: depth alignment module}
\begin{gathered}
    \mathbf{A}_i, \mathbf{B}_i= f_{\text{LWLR}}(\mathbf{D}^g_i, \{\omega_{i, t} \cdot  d^{g}_i ( {\mathbf{p}_{t}} ) \}_{t=1}^M )  \\
    \mathbf{D}_i = \mathbf{A}_i \odot \mathbf{D}^g_i + \mathbf{B}_i
\end{gathered}
\end{equation}
where $\odot$ means element-wise multiplication, and $M$ is set to 25 in our experiments.
Therefore, in this module, we only need to optimize 
for 
$27$ parameters for each frame, \textit{i.e.}, $\alpha_i, \beta_i, \bm{\omega}_{i}$. With our geometric consistency alignment module, we can achieve scale-consistent depth.

\noindent\textbf{Representation of Camera Poses and Intrinsic Camera Parameters.}
We propose to optimize
for 
the relative rotation vector $\{\mathbf{r}_{i\rightarrow (i+1)}\}_{i=1}^{P-1}$ (\textit{i.e.},  Euler angles) and relative translation vector $\{\mathbf{t}_{i\rightarrow (i+1)}\}_{i=1}^{P-1}$ between two adjacent frames. They are initialized to 0 and are transformed to $4 \times 4$ relative pose matrices $\{\mathbf{P}_{i\rightarrow (i+1)}\}_{i=1}^{P-1}$ with $\mathcal{H}(\cdot, \cdot )$. 
The relative pose matrices $\{\mathbf{P}_{i\rightarrow (i+1)}\}_{i=1}^{P-1}$ are transformed into camera-to-world poses matrices $\{\mathbf{P}_{i}\}_{i=1}^{P}$ with the product operation. Such initialization ensures the initial overlap of keyframes. 
For intrinsic camera parameters, we assume a simple pinhole camera model, initialize the focal length to be $f_{0} = 1.2 \cdot \max(H, W)$, and 
learn the focal length with a learnable scalar $\delta$. We assume 
that 
the optical center is at the image center. Thus for camera parameters, we need to optimize 
for 
$6(P-1) + 1$ %
variables: 
\begin{equation}
\label{eq: pose_and_intrinsic}
\begin{gathered}
    \mathbf{P}_{i\rightarrow (i+1)} = \mathcal{H}(\mathbf{r}_{i\rightarrow (i+1)}, \mathbf{t}_{i\rightarrow (i+1)}), \;\;\;
    \mathbf{P}_{1} = \mathbf{E}_{4,4}, \\
    \mathbf{P}_{i} = \mathbf{P}_{1} \prod_{k=1}^{i-1} \mathbf{P}_{k\rightarrow (k+1)}, 
    \;\;\;
    \mathbf{K} = \begin{bmatrix} \delta f_{0} & 0 &  \nicefrac{W}{2} 
    \\ 0 & \delta f_{0} & 
    \nicefrac{H}{2}
    \\ 0 & 0 & 1  \end{bmatrix}
\end{gathered}
%\vspace{-0.5 em}
\end{equation}
where 
$\mathbf{E}_{4,4}$ is the $4\times4$ elementary matrix.

\noindent\textbf{Optimization Objectives.} %
The image coordinate $\mathbf{p}_{i}$ in frame $i$ %
can be %
wrapped to frame $j$ as follows. 
\def\R{{\cal R}}
\begin{equation}
\label{eq: unprojection}
\begin{gathered}
    f_{\text{s}}(\mathbf{D}_{i\rightarrow j}, \mathbf{p}_i) \mathbf{p}_{i\rightarrow j} = \mathbf{K} \mathbf{R}_j^\mathsf{T} 
    \bigl[\mathbf{R}_i \mathbf{K}^{-1} f_{\text{s}}(\mathbf{D}_i, \mathbf{p}_i) \mathbf{p}_i + \mathbf{t}_i - \mathbf{t}_j
    \bigr], \\
    \mathbf{p}_i, \mathbf{p}_{i\rightarrow j}, \mathbf{t}_i, \mathbf{t}_j \in \mathbb{R}^{3 \times 1}, \;\;\; \mathbf{R}_i, \mathbf{R}_j, \mathbf{K} \in \mathbb{R}^{3 \times 3}
\end{gathered}
\end{equation}
where $\mathbf{K}$ is the intrinsic camera parameter. $\mathbf{R}_i$ and $\mathbf{t}_i$ are the camera-to-world rotation and translation matrices, and $\mathbf{P}_i = \begin{bmatrix} \mathbf{R}_i & \mathbf{t}_i
    \\ \mathbf{0} & 1 \end{bmatrix}$. %
$\mathbf{p}_{i\rightarrow j}$ is the %
point warped from %
$\mathbf{p}_{i}$ of frame $i$ to frame $j$, and $\mathbf{D}_{i\rightarrow j}$ is the %warp computed 
warped 
depth map.

To optimize the proposed variables including $\{\alpha_i\}_{i=1}^P$, $\{\beta_i\}_{i=1}^P$, $\delta$, $\{\mathbf{r}_{i\rightarrow (i+1)}\}_{i=1}^{P-1}$, $\{\mathbf{t}_{i\rightarrow (i+1)}\}_{i=1}^{P-1}$, and $\{\bm{\omega}_i\}_{i=1}^{P}$, we propose to use the pixel-wise photometric and geometric constraint  \cite{bian2019unsupervised} together to ensure the consistency of color and depth along the video as follows.
\begin{eqnarray}
 L_{pc} = \frac{1}{\left| V\right|} \sum_{\mathbf{p} \in V \atop (i, j) \in K}
    \lvert f_{\text{s}}(\mathbf{I}_{i}, \mathbf{p}_{i}) - f_{\text{s}}(\mathbf{I}_{j}, \mathbf{p}_{i\rightarrow j}) \rvert
\label{eq: photometric constraint}
\end{eqnarray}
%\vspace{-1 em}

\begin{eqnarray}
  L_{gc} = \frac{1}{\left| V\right|} \sum_{\mathbf{p} \in V \atop (i, j) \in K}
  \frac{\left| f_{\text{s}}(\mathbf{D}_{j}, \mathbf{p}_{i\rightarrow j}) - f_{\text{s}}(\mathbf{D}_{i\rightarrow j}, \mathbf{p}_{i}) \right|}{f_{\text{s}}(\mathbf{D}_{j}, \mathbf{p}_{i\rightarrow j}) + f_{\text{s}}(\mathbf{D}_{i\rightarrow j}, \mathbf{p}_{i})}
\label{eq: geometric constraint}
\end{eqnarray}
where $V$ represents all valid points successfully projected from frame $i$ to frame $j$, $K$ is the selected keyframe pairs. 

Furthermore, we enforce a regularization term on the sampled sparse anchor points:
\begin{equation}
\label{eq: regularization}
\begin{gathered}
   L_{regu} = \sum_{i=1}^N \sum_{t=1}^M \left| 1- \omega_{i, t} \right|
\end{gathered}
\end{equation}

The overall constraints are as follows.
\begin{equation}
\label{eq: loss_function}
\begin{gathered}
   L =  \lambda _{pc} L_{pc} + \lambda_{gc} L_{gc} + \lambda _{regu}L_{regu}
\end{gathered}
\end{equation}
where $\lambda _{pc}$, $\lambda _{gc}$, and $\lambda _{regu}$ are weights to balance them.

\noindent\textbf{Optimization Details.}
Before we perform optimization, we
first sample $P$ frames from the video to reduce the time complexity. With these sampled frames, we perform the keyframes sampling and optimization, which consists of two stages, \textit{i.e.}, local keyframes sampling for optimization, and global keyframes sampling for optimization. %

\textit{Local keyframes sampling and optimization.}
In optimization, we first sample local keyframes from the  $k$ 
nearest 
frames of each reference frame.
$k$ is set to $6$ in our experiments. To reduce computation time complexity, we do not
use 
all these local keyframes with the current frame but select them based on probability.  %
The probability for each keyframe is set as follows. 
\begin{equation}
\label{eq: sample_prob_stage_one}
\begin{gathered}
	p_{l} =   \left\{
	            \begin{aligned}
	            \frac{1}{k},& \,\,{\rm  if} \,k\text{-}{\rm nearest ~ neighbors}, \\
	            0,& ~~%
             {\rm otherwise},
	            \end{aligned}
            \right.
\end{gathered}
\end{equation}
where $p_{l}$ is the keyframe sample probability.

\begin{algorithm}[t]
\algsetup{linenosize=\footnotesize} %
\small 
\renewcommand{\algorithmicrequire}{\textbf{Input:}}
\renewcommand{\algorithmicensure}{\textbf{Output:}}
    \caption{Optimization Algorithm}
    \label{alg: optimization}
    \begin{algorithmic}[1]
        \REQUIRE $ N $  images $\{\mathbf{I}_{i}\}_{i=1}^N$\\
        \ENSURE scale-consistent depth maps $\{\mathbf{D}_{i}\}_{i=1}^P$, intrinsic camera parameter $\mathbf{K}$, camera-to-world camera poses $\{\mathbf{P}_i\}_{i=1}^{P-1}$ \\

        \STATE Sample $P$ images $\{\mathbf{I}_{i}\}_{i=1}^P$ from $\{\mathbf{I}_{i}\}_{i=1}^N$
        \STATE %
        Obtain affine-invariant depth $\{\mathbf{D}^a_{i}\}_{i=1}^P$ from $\{\mathbf{I}_{i}\}_{i=1}^P$
        \STATE Initialize the global scale $\{\alpha_i\}_{i=1}^P$, global shift $\{\beta_i\}_{i=1}^P$, focal length scalar $\delta$, relative rotation $\{\mathbf{r}_{i\rightarrow (i+1)}\}_{i=1}^{P-1}$, relative translation $\{\mathbf{t}_{i\rightarrow (i+1)}\}_{i=1}^{P-1}$, sparse points weight $\{\bm{\omega}_i\}_{i=1}^{P}$

        \FOR{$stage$ in [$Local$, $Global$]}
            \FOR{$iter$ = 1 to $iterations$}
                \STATE Compute $\{\mathbf{P}_i\}_{i=1}^{P-1}$ and $\mathbf{K}$ with Eq.~\eqref{eq: pose_and_intrinsic} \\
                \IF{$stage$ = $Local$}
                    \STATE Sample keyframe pairs $\{(i_k, j_k)\}_{k=1}^K$ with Eq.~\eqref{eq: sample_prob_stage_one}
                \ELSE
                    \STATE Sample keyframe pairs $\{(i_k, j_k)\}_{k=1}^K$ with Eq.~\eqref{eq: sample_prob_stage_two}
                \ENDIF
            
                \FOR{keyframe pair $(i, j)$ in $\{(i_k, j_k)\}_{k=1}^K$}
                    \STATE Compute scale-consistent depth $\mathbf{D}_i$, $\mathbf{D}_j$ with Eq.~\eqref{eq: depth alignment module}; \\
                    
                    \STATE Warp $\mathbf{p}_i$ from frame $i$ to frame $j$, get warped depth $\mathbf{D}_{i\rightarrow j}$ and warped locations $\mathbf{p}_{i\rightarrow j}$ with Eq.~\eqref{eq: unprojection}

                    \STATE Compute optimization objectives with Eq.~\eqref{eq: loss_function}
                \ENDFOR
                
                \STATE Back-propagate and update $\{\alpha_i\}_{i=1}^P$, $\{\beta_i\}_{i=1}^P$, $\delta$, $\{\mathbf{r}_{i\rightarrow (i+1)}\}_{i=1}^{P-1}$, $\{\mathbf{t}_{i\rightarrow (i+1)}\}_{i=1}^{P-1}$, $\{\bm{\omega}_i\}_{i=1}^{P}$
            \ENDFOR
        \ENDFOR
        \STATE Compute scale-consistent depth maps $\{\mathbf{D}_{i}\}_{i=1}^P$ with Eq.~\eqref{eq: depth alignment module}, compute camera parameters $\mathbf{K}$ and $\{\mathbf{P}_i\}_{i=1}^{P-1}$ with Eq.~\eqref{eq: pose_and_intrinsic}
    \end{algorithmic}
\end{algorithm}

\begin{table*}[t]
\newcommand{\tabincell}[2]{\begin{tabular}{@{}#1@{}}#2\end{tabular}}
  \centering
  \caption{\textbf{Quantitative comparison of zero-shot 3D scene reconstruction with state-of-the-art methods.} 
  We compare with seven categories of reconstruction algorithms on several video sequences of five unseen datasets: NYU  \cite{silberman2012indoor}, ScanNet  \cite{dai2017scannet}, 7-Scenes  \cite{shotton2013scene}, TUM  \cite{sturm2012benchmark}, and KITTI  \cite{geiger2013vision}. Note that NeuralRecon is trained on ScanNet  \cite{dai2017scannet}, 
  $^*$ denotes ground-truth camera poses are given, and ``Rank" means the average ranking performance of each column. We evaluate the Chamfer distance C-$l_1$ and the F-score with a threshold of 5cm. As a result, our method achieves state-of-the-art rank on the five zero-shot datasets.}
 \vspace{-0.5 em}
\resizebox{\linewidth}{!}{%
  \begin{tabular}{@{}r|lr|lr|lr|lr|lr|l@{}}
    \toprule
	
	\multirow{2}{*}{Method} & \multicolumn{2}{c|}{NYU  \cite{silberman2012indoor}} & \multicolumn{2}{c|}{ScanNet  \cite{dai2017scannet}} & \multicolumn{2}{c|}{7-Scenes  \cite{shotton2013scene}} & \multicolumn{2}{c|}{TUM  \cite{sturm2012benchmark}} & \multicolumn{2}{c|}{KITTI  \cite{geiger2013vision}} & \multirow{2}{*}{Rank}\\
	
	\cline{2-11}
	
    & C-$l_1$ $\downarrow$ & F-score$\uparrow$ & C-$l_1$ $\downarrow$ & F-score$\uparrow$ & C-$l_1$ $\downarrow$ & F-score$\uparrow$ & C-$l_1$ $\downarrow$ & F-score$\uparrow$ & C-$l_1$ $\downarrow$ & F-score$\uparrow$ \\
    
    \hline

    NeuralRecon$^*$  \cite{sun2021neuralrecon}& 0.487 & 0.185
    			& \multicolumn{2}{c|}{trained on ScanNet}
    			& 0.235 & 0.219 
    			& 0.435 & 0.241
                    & \multicolumn{2}{c|}{failed on KITTI}
    			& 8.125
    			 \\

    \hline

    DPSNet$^*$  \cite{im2018dpsnet}	& 0.198 & 0.391
    			& 0.299 & 0.266
    			& 0.574 & 0.142 
    			& 0.336 & 0.263
                    & \textbf{0.290} & \textbf{0.232}
    			& 4.800
    			 \\
    
    \hline
    BoostingDepth-DROID  \cite{xu2022boosting}	& 0.139 & 0.481
    			& 0.379 & 0.292
    			& 0.235 & 0.460
    			& 0.322 & 0.439
                    & 2.431 & 0.010
    			& 3.600
    			 \\

    \hline

    SC-DepthV3  \cite{sun2022sc}	& 0.196 & 0.458
    			& 0.402 & 0.214
    			& 0.252 & 0.240
    			& 0.525 & 0.244
                    & 4.133 & 0.036
    			& 5.500
    			 \\
    
    \hline
    
    CVD  \cite{luo2020consistent} 	& 0.471 & 0.302
    		& \multicolumn{2}{c|}{failed on ScanNet}
    		& 0.416 & 0.215  
    		& 0.378 & 0.239
                & 5.479 & 0.029
    		& 7.900
    		\\
    
    RCVD  \cite{kopf2021rcvd} 	& 0.303 & 0.346
    		& 0.641	& 0.125
    		& 0.497 & 0.182
    		& 0.679 & 0.218
                & 58.372 & 0.020
    		& 8.300
			\\
    
    GCVD  \cite{lee2022globally} 	& 0.148 & 0.453
    		& 0.631 & 0.147
    		& 0.196 & 0.326
    		& 0.350 & 0.339
                & 2.127 & 0.114
    		& 4.400
    		\\
    
    \hline

    COLMAP  \cite{schonberger2016structure, schonberger2016pixelwise}  & 0.251 & 0.343
            & 0.796 & 0.127
            & 0.513 & 0.178
            & 0.385 & 0.249
            & 107.451 & 0.152
            & 7.300
    			\\
    \hline

    DROID-SLAM  \cite{teed2021droid} 	& 0.224 & 0.516
    			& 0.416 & 0.384
    			& 0.304 & \textbf{0.469}
    			& 0.285 & 0.433
                    & 0.686 & 0.155
    			& 3.200
    			\\
     
     \hline
     
     Ours 	& \textbf{0.099} & \textbf{0.622}
     		& \textbf{0.170} & \textbf{0.410} 
     		& \textbf{0.170} & 0.464
     		& \textbf{0.211} & \textbf{0.453}
                & 0.670 & 0.151
     		& \textbf{1.500}
     		\\
     
    \bottomrule
  \end{tabular}}
%  \vspace{-1 em}
  \label{tab: recon_comparison}
\end{table*}

\begin{table*}[t]
\newcommand{\tabincell}[2]{\begin{tabular}{@{}#1@{}}#2\end{tabular}}
  \centering
  \caption{\textbf{Quantitative comparison of zero-shot depth estimation with state-of-the-art methods.} We evaluate the absolute relative error (AbsRel) and percentage of accurate depth pixels ($\delta_1$) on several videos of five unseen datasets. ``Rank" means the average ranking performance of each column. The gray ``\gray{LeReS}" represents the performance of affine-invariant depth without any alignment with ground-truth depth. As a result, our algorithm achieves state-of-the-art rank performance.}
\vspace{-0.5 em}
\resizebox{\linewidth}{!}{%
  \begin{tabular}{@{}r|lr|lr|lr|lr|lr|l@{}}
    \toprule
	
	\multirow{2}{*}{Method} & \multicolumn{2}{c|}{NYU  \cite{silberman2012indoor}} & \multicolumn{2}{c|}{ScanNet  \cite{dai2017scannet}} & \multicolumn{2}{c|}{7-Scenes  \cite{shotton2013scene}} & \multicolumn{2}{c|}{TUM  \cite{sturm2012benchmark}} & \multicolumn{2}{c|}{KITTI  \cite{geiger2013vision}} & \multirow{2}{*}{Rank}\\
	
	\cline{2-11}
	
    & AbsRel$\downarrow$ & $\delta_1$$\uparrow$ & AbsRel$\downarrow$ & $\delta_1$$\uparrow$ & AbsRel$\downarrow$ & $\delta_1$%
$\uparrow$ & AbsRel$\downarrow$ & $\delta_1$$\uparrow$ & AbsRel$\downarrow$ & $\delta_1$$\uparrow$ \\

    \hline

    DPSNet{$^*$}  \cite{im2018dpsnet}	& 0.200 & 0.662
    			& 0.216 & 0.630
    			& 0.201 & 0.657 
    			& 0.331 & 0.457
                    & 0.188 & 0.729 
    			& 6.4
    			 \\

    \hline
    BoostingDepth-DROID  \cite{xu2022boosting}	& 0.096 & 0.921
    			& 0.236 & 0.685
    			& 0.145 & 0.828
    			& 0.180 & 0.742
                    & \textbf{0.112} & 0.860
    			& 2.7
    			 \\
    
    \hline
    
    SC-DepthV3  \cite{sun2022sc}	& 0.120 & 0.858  
    			& 0.197 & 0.672
    			& 0.188 & 0.688
    			& 0.254 & 0.569 
                    & 0.287 & 0.463
    			& 5.2
    			 \\
    
    \hline
    
    CVD  \cite{luo2020consistent} 	& 0.167	& 0.837	
     		& \multicolumn{2}{c|}{failed on ScanNet}
     		& 0.180	& \textbf{0.847}	
     		& 0.148	& 0.797
                & 0.717 & 0.215 
     		& 5.6
    		\\
    
    RCVD  \cite{kopf2021rcvd} 	& 0.192	& 0.700	
     		& 0.235	& 0.620	
     		& 0.241	& 0.635	
     		& 0.246	& 0.612	
                & 0.220 & 0.606 
     		& 6.6
			\\
    
    GCVD  \cite{lee2022globally} 	& 0.163	& 0.768	
     		& 0.316	& 0.527	
     		& 0.203	& 0.650	
     		& 0.248	& 0.697	
                & 0.240 & 0.543 
     		& 6.4
    		\\
    
    \hline
    
    COLMAP  \cite{schonberger2016pixelwise, schonberger2016structure}  & 0.233 & 0.722
                & 0.562 & 0.409
                & 0.268 & 0.666
                & 0.269 & 0.665
                & 0.302 & 0.773
                & 7.2
            \\
    \hline
    
    DROID-SLAM  \cite{teed2021droid} 	& 0.143	& 0.847
     			& 0.210	& 0.773	
     			& 0.154	& 0.827	
     			& 0.210	& 0.720	
                    & 0.115 & \textbf{0.914} 
     			& 3.1
     			\\
     
     \hline

     \gray{LeReS}   \cite{yin2021learning} & \gray{0.277} & \gray{0.508} 
                                        & \gray{0.409} & \gray{0.339} 
                                        & \gray{0.414} & \gray{0.345} 
                                        & \gray{0.445} & \gray{0.320} 
                                        & \gray{0.298} & \gray{0.439} 
                                        & \gray{-} 
                                        \\
     
     Ours 	& \textbf{0.092}	& \textbf{0.923}
     		& \textbf{0.127}	& \textbf{0.858}	
     		& \textbf{0.135}	& 0.844
     		& \textbf{0.145}	& \textbf{0.799}
                & 0.203 & 0.632 
     		& \textbf{1.8}
     		\\
     
    \bottomrule
  \end{tabular}}
%  \vspace{-1 em}
  \label{tab: depth_comparison}
\end{table*}

\begin{table*}[t]
\newcommand{\tabincell}[2]{\begin{tabular}{@{}#1@{}}#2\end{tabular}}
  \centering
  \caption{\textbf{Quantitative comparison of zero-shot pose estimation.} ``ATE'', ``RPE-T'', and ``RPE-R'' stand for the absolute trajectory error and the relative pose error of translation and rotation consecutively. ``\%'' represents the percentage of successfully tracked frames. Note that ORB-SLAM3 fails to track the trajectory on some low-texture regions, and can only estimate partial camera poses. Our algorithm achieves comparable performance to ORB-SLAM3 on three unseen datasets and better performance than the video depth estimation algorithms.
  }
  % \vspace{-1 em}
 \resizebox{\linewidth}{!}{%
 \setlength{\tabcolsep}{0.5mm}
\small 
  \begin{tabular}{@{}r|c|lccr|lccr|lccr|lccr|lccr@{}}
    \toprule
	
	\multirow{2}{*}{Method} & GT & \multicolumn{4}{c|}{NYU  \cite{silberman2012indoor}} & \multicolumn{4}{c|}{ScanNet  \cite{dai2017scannet}} & \multicolumn{4}{c|}{7-Scenes  \cite{shotton2013scene}} & \multicolumn{4}{c|}{TUM  \cite{sturm2012benchmark}} & \multicolumn{4}{c}{KITTI  \cite{geiger2013vision}}\\
	
	\cline{3-22}
 
    & Intrinsic & ATE$\downarrow$ & RPE-T$\downarrow$ & RPE-R$\downarrow$ & \%$\uparrow$
    & ATE$\downarrow$ & RPE-T$\downarrow$ & RPE-R$\downarrow$ & \%$\uparrow$
    & ATE$\downarrow$ & RPE-T$\downarrow$ & RPE-R$\downarrow$ & \%$\uparrow$
    & ATE$\downarrow$ & RPE-T$\downarrow$ & RPE-R$\downarrow$ & \%$\uparrow$
    & ATE$\downarrow$ & RPE-T$\downarrow$ & RPE-R$\downarrow$ & \%$\uparrow$\\
    
    \hline

    SC-DepthV3  \cite{sun2022sc} &  \checkmark     & 0.241 & 0.444 & 0.115 & 1.000
                            			& 0.492 & 0.702 & 0.327 & 1.000
                                            & 1.011 & 1.644 & 0.389 & 1.000
                                            & 0.528 & 0.787 & 0.399 & 1.000
                                            & 1.796 & 4.186 & 0.077 & 1.000
                            			\\
    \hline

    CVD  \cite{luo2020consistent} &  \checkmark     & 1.192 & 2.049 & 0.107 & 1.000
                            			& 0.734 & 1.197 & 0.257 & 1.000
                                            & 0.532 & 1.644 & 0.378 & 1.000
                                            & 0.268 & 0.844 & 0.447 & 1.000
                                            & 19.32 & 24.50 & 0.027 & 1.000 
                            			\\

    RCVD  \cite{kopf2021rcvd} &      & 0.460 & 1.263 & 0.114 & 1.000
                            			& 1.025 & 1.708 & 0.615 & 1.000
                                            & 0.559 & 1.545 & 0.394 & 1.000
                                            & 0.626 & 1.083 & 0.421 & 1.000
                                            & 134.3 & 160.7 & 0.463 & 1.000 
                            			\\

    GCVD  \cite{lee2022globally} &     & 0.160 & 0.237 & 0.083 & 1.000
                            			& 0.573 & 0.979 & 0.620 & 1.000
                                            & 0.259 & 0.368 & 0.162 & 1.000
                                            & 0.162 & 0.204 & 0.113 & 1.000
                                            & 5.678 & 8.733 & 0.089 & 1.000 
                            			\\
    \hline

    COLMAP  \cite{schonberger2016pixelwise, schonberger2016structure}  &   & 0.091 & 0.137 & 0.055 & 1.000
                & 0.352 & 0.426 & 0.096 & 1.000
                & 0.062 & 0.090 & 0.051 & 1.000
                & 0.075 & 0.102 & 0.064 & 1.000
                & 3.707 & 3.454 & 0.016 & 1.000  \\
    
    \hline
    
    DROID-SLAM  \cite{teed2021droid} &  \checkmark     & 0.050 & 0.079 & 0.032 & 1.000
                            			& 0.230 & 0.230 & 0.052 & 1.000
                            			& 0.050 & 0.072 & 0.072 & 1.000
                            			& 0.044 & 0.077 & 0.079 & 1.000
                                            & 1.491 & 2.302 & 0.005 & 1.000
                            			\\

    ORB-SLAM3  \cite{campos2021orb} & \checkmark	    & 0.065 & 0.097 & 0.039 & 0.791
                        			& 0.208 & 0.300 & 0.098 & 0.426
                        			& 0.118 & 0.211 & 0.121 & 0.956
                        			& 0.104 & 0.144 & 0.082 & 0.504
                                        & 3.124 & 3.926 & 0.009 & 0.971
                        			\\
     
     \hline

    Ours             &  & 0.079 & 0.121 & 0.046 & 1.000
                        & 0.130 & 0.184 & 0.085 & 1.000
                        & 0.141 & 0.229 & 0.075 & 1.000
                        & 0.176 & 0.228 & 0.151 & 1.000
                        & 3.541 & 5.663 & 0.053 &  1.000
                        \\
     
    \bottomrule
  \end{tabular}
 }
  \label{tab: pose_comparison}
  % \vspace{-1 em}
\end{table*}

\begin{table}[h]
\newcommand{\tabincell}[2]{\begin{tabular}{@{}#1@{}}#2\end{tabular}}
  \centering
  \caption{\textbf{Quantitative comparison of zero-shot intrinsic camera parameter estimation.} ``FOV AbsRel'' represents the absolute relative error of the field of view. Note that only three algorithms can estimate the intrinsic camera parameter  from the input video, and our algorithm can achieve accurate and robust performance on four unseen datasets. 
  }
  % \vspace{-1 em}
 \resizebox{\linewidth}{!}{%
 \setlength{\tabcolsep}{0.5mm}
  \begin{tabular}{@{}r|c|c|c|c|l@{}}
    \toprule
	
	\multirow{2}{*}{Method} & \multicolumn{1}{c|}{NYU  \cite{silberman2012indoor}} & \multicolumn{1}{c|}{ScanNet  \cite{dai2017scannet}} & \multicolumn{1}{c|}{7-Scenes  \cite{shotton2013scene}} & \multicolumn{1}{c|}{TUM  \cite{sturm2012benchmark}} & \multirow{2}{*}{Rank}\\
	
	\cline{2-5}
 
    & FOV AbsRel$\downarrow$ & FOV AbsRel$\downarrow$ & FOV AbsRel$\downarrow$ & FOV AbsRel$\downarrow$ &   \\
    
    \hline

    RCVD  \cite{kopf2021rcvd}   & 0.144  & 0.148 & 0.110 & 0.174 & 2.75 \\

    GCVD  \cite{lee2022globally}   & 0.015  & 0.126 & 0.143 & 0.043 & 1.75 \\
     
     \hline

    Ours   & 0.032  & 0.032 & 0.085 & 0.056 & 1.50 \\
     
    \bottomrule
  \end{tabular}
 }
  \label{tab: intrinsic_comparison}
  % \vspace{-1 em}
\end{table}

\textit{Global keyframes sampling and optimization.} %
We further perform global keyframes sampling and optimization to improve long-range consistency. For each reference frame, all other frames are regarded as paired keyframes, but they are labeled with different sampling probabilities. We compute the relative pose between each reference frame $\mathbf{I}_i$ and any other frame $\mathbf{I}_j$. Based on the relative pose angle $\theta_{ij}$, we set the global sampling probability $p_g$ as follows. %
\begin{equation}
\label{eq: sample_prob_stage_two}
\begin{gathered}
	p_{g} = \frac{p_{l} + p}{2},p =   \left\{
	            \begin{aligned}
	             \frac{\theta_{ij}}{\phi^2}, & \;\; {\rm if} \;0<\theta_{ij}\leq\phi   \\
	             \frac{2}{\phi} - \frac{\theta_{ij}}{\phi^2},&\;\; {\rm if} \;\phi<\theta_{ij}<2\phi  \\
	             0,&\;\; {\rm otherwise}  \\
	            \end{aligned}
            \right.
\end{gathered}
\end{equation}
where $\phi$ is an angle threshold, and $\theta_{ij}$ is the relative rotation angle between the source frame $\mathbf{I}_i$ and the reference frame $\mathbf{I}_j$. $p_l$ is the local probability with Eq.~\eqref{eq: sample_prob_stage_one}. The optimization is summarized in Algorithm~\ref{alg: optimization}.

\section{Experiments}
\label{sec:experiments}

We evaluate the 3D reconstruction, scale-consistent depths, optimized camera poses, and optimized intrinsic camera parameters in experiments. More details and analyses including runtime can be found in the supplementary.

\subsection{Dense 3D Scene Reconstruction}
To %
show the robustness and accuracy of our reconstruction %
method, we compare it with a learning-based volumetric 3D reconstruction method (NeuralRecon  \cite{sun2021neuralrecon}), a multi-view depth estimation method (DPSNet  \cite{im2018dpsnet}), a per-frame scale-shift alignment method (BoostingDepth  \cite{xu2022boosting}), an %
unsupervised video depth estimation method (SC-DepthV3  \cite{sun2022sc}), some %
consistent video depth estimation methods (CVD  \cite{luo2020consistent}, RCVD  \cite{kopf2021rcvd}, GCVD  \cite{lee2022globally}), a Structure-from-Motion method COLMAP  \cite{schonberger2016pixelwise, schonberger2016structure}, and a robust deep visual SLAM method (DROID-SLAM  \cite{teed2021droid}) on several video sequences of $5$ zero-shot datasets, \ie, NYU  \cite{silberman2012indoor}, ScanNet  \cite{dai2017scannet}, 7-Scenes  \cite{shotton2013scene}, TUM  \cite{sturm2012benchmark}, KITTI  \cite{geiger2013vision}. 
Note that NeuralRecon is trained on ScanNet, while others are evaluated on unseen scenarios. NeuralRecon and DPSNet %
require GT poses and camera intrinsic as input, and SC-DepthV3, CVD, and DROID-SLAM require camera intrinsic. For fair comparison, we employ the estimated poses and depths of DROID-SLAM for BoostingDepth (BoostingDepth-DROID) to iteratively align the scale-shift values and filter outliers, as their paper does. Other methods %
optimize camera parameters on their own. 

Quantitative comparisons of zero-shot 3D reconstruction and depth %
are shown in Table~\ref{tab: recon_comparison} and Table \ref{tab: depth_comparison} respectively. %
NeuralRecon  \cite{sun2021neuralrecon}, DPSNet  \cite{im2018dpsnet}, SC-DepthV3  \cite{sun2022sc} need to optimize millions of parameters, thus showing less robustness to unseen datasets. 
Rather than purely per-frame aligning with offline-obtained depth after filtering as BoostingDepth~  \cite{xu2022boosting} does, we jointly optimize the camera parameters and depth parameters, which ensures consistency between frames and less suffers from outliers of sparse points. 
CVD  \cite{luo2020consistent} fails to %
reconstruct half of the ScanNet scenarios. %
Compared with RCVD  \cite{kopf2021rcvd}, and GCVD  \cite{lee2022globally}, our %
method employs a global and a local scale-shift recovery strategy, which can rectify the affine-invariant depth to a much more accurate scale-consistent depth. Thus, we can achieve much more accurate reconstructions.
The COLMAP  \cite{schonberger2016pixelwise, schonberger2016structure} struggles to estimate dense depth maps, and thus results in sparse 3D reconstruction. 
The robust DROID-SLAM  \cite{teed2021droid}
focuses on optimizing more robust poses and trajectories with the recurrent module, and spends less effort on dense reconstruction. Thus, their dense reconstruction 
is more likely to 
suffer from %
outliers. In contrast, our optimization is based on dense depth priors and employs a global and local optimization strategy to reduce noise, thus we %
in general 
achieve better reconstruction performance. 

\begin{table}[t]
\newcommand{\tabincell}[2]{\begin{tabular}{@{}#1@{}}#2\end{tabular}}
\caption{\textbf{Runtime analysis on three representative scenes.} Our pipeline achieves state-of-the-art reconstruction but only takes around a quarter of an hour to optimize.}
  \centering
  % \vspace{-1 em}
 \resizebox{\linewidth}{!}{%
 \setlength{\tabcolsep}{0.5mm}
\small 
  \begin{tabular}{@{}r|c|c|c@{}}
    \toprule
	
	Method & basement\_0001a & bedroom\_0015 & chess \\
	
	\hline

    NeuralRecon\cite{sun2021neuralrecon} & - & - & - \\
    DPSNet\cite{im2018dpsnet} & 5m 53s & 45s & 12m 20s \\
    BoostingDepth-DROID\cite{xu2022boosting} & 35s & 10s & 47s \\
    SC-DepthV3\cite{sun2022sc} & 1m 47s & 17s & 1m 54s \\
    CVD\cite{luo2020consistent} & 1h 7m 6s & 12m 30s & 8h 1m 34s \\
    RCVD\cite{kopf2021rcvd} & 1h 15m 45s & 10m 6s & 6h 14m 54s \\
    GCVD\cite{lee2022globally} & 10m 26s & 2m 37s & 52m 15s \\
    DROID-SLAM\cite{teed2021droid} & 35s & 16s & 1m 18s \\
    % ORB-SLAM3 & 11s & 6s & 38s \\
    COLMAP\cite{schonberger2016structure, schonberger2016pixelwise} & 50m 44s & 8m 34s & 9h 40m 49s \\
    Ours & 14m 55s & 6m 55s & 21m 8s \\
    \bottomrule
  \end{tabular}
 }
  % \vspace{-1.5 em}
  \label{tab: runtime analysis}
\end{table}

Quantitative comparisons of zero-shot pose estimation and intrinsic camera parameter estimation are shown in Table~\ref{tab: pose_comparison} and Table~\ref{tab: intrinsic_comparison}. Our algorithm achieves comparable performance with the traditional SLAM methods ORB-SLAM3  \cite{campos2021orb} on three datasets. However, the ORB-SLAM3 can only estimate the camera poses of partial frames due to the failure of tracking features. In contrast, our algorithm can optimize and interpolate to obtain dense trajectories. Compared to four video depth estimation methods, we can outperform them by a large margin. Note that only ours, RCVD, and GCVD can optimize the intrinsic camera parameters, while others can only employ GT camera intrinsic as input. For intrinsic camera parameter estimation, our algorithm can achieve robust and accurate performance on four unseen datasets.

\begin{figure*}[t]
\centering    %
\includegraphics[width=\linewidth]{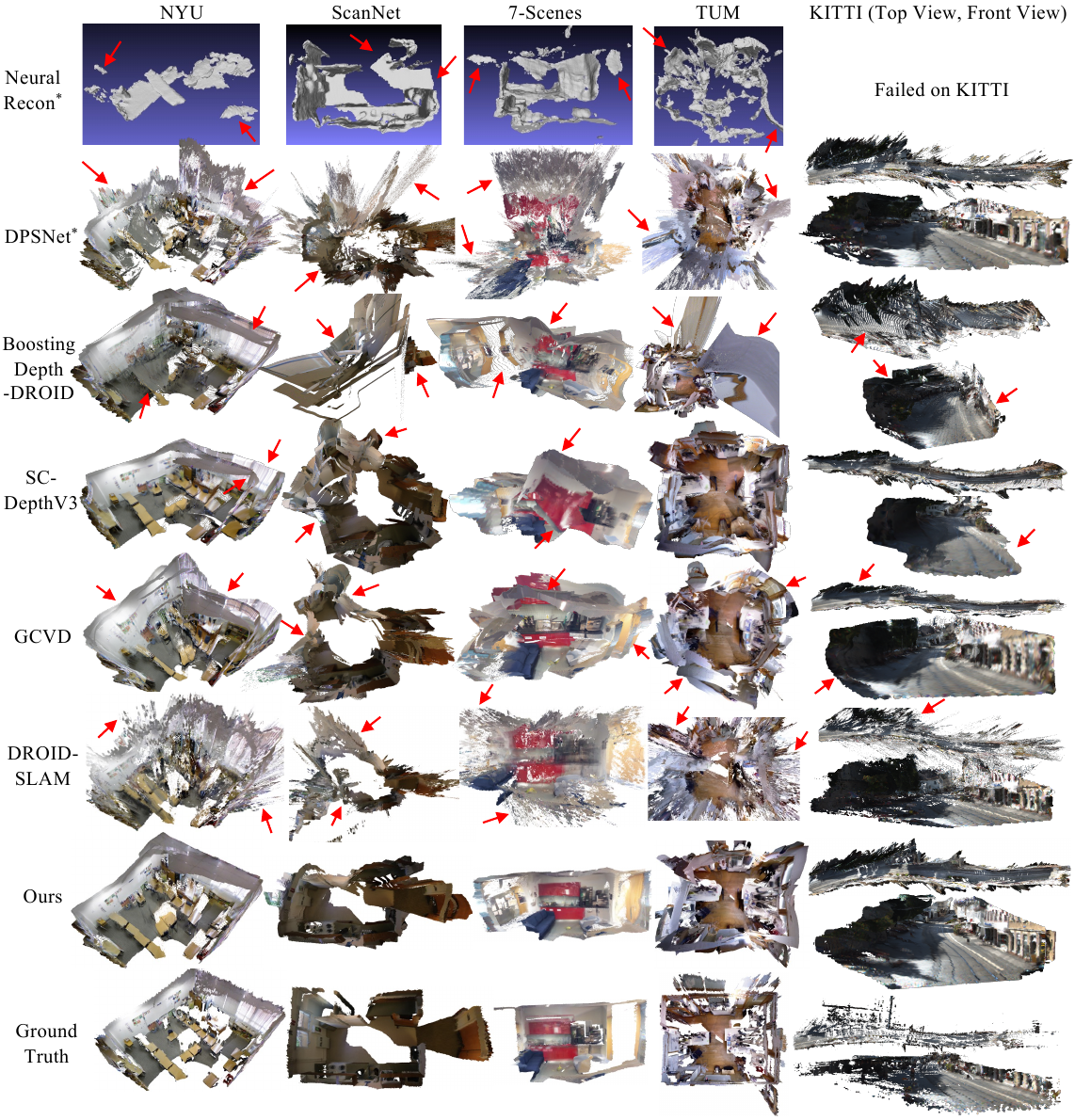}
% \vspace{-1 em}
\caption{\textbf{Qualitative comparison of zero-shot 3D scene reconstruction from a monocular video.} 
We compare with five representative algorithms on five unseen datasets. 
Note that NeuralRecon is trained on ScanNet  \cite{dai2017scannet} and can only output uncolored mesh, and $^*$ represents the employment of ground-truth camera poses during reconstruction. These methods either suffer from inaccurate pose estimation, or predict noise depth maps (please see the red arrows). As can be seen, our algorithm can reconstruct better 3D scene shape without any offline-computed camera parameters.
} %
% \vspace{-1 em}
\label{fig: 3d_recon}
\end{figure*}

Qualitative comparisons %
are
reported 
in Figure.~\ref{fig: 3d_recon}. %
NeuralRecon can only reconstruct partial meshes of unseen scenarios. The DPSNet  \cite{im2018dpsnet} suffers from inaccurate feature correspondences, which %
cause noisy %
point clouds. BoostingDepth  \cite{xu2022boosting} can only filter out inaccurate points of DROID-SLAM but cannot ensure the temporal consistency, and thus fails to reconstruct consistent mesh. The SC-DepthV3  \cite{sun2022sc} suffers from the weak supervision of photometric loss and %
work worse on some low-texture regions. %
Consistent video depth estimation methods such as GCVD \cite{lee2022globally} can not recover the shift of depth maps, which %
will cause distortions in 3D point clouds. DROID-SLAM  \cite{teed2021droid} concentrates on accurate pose estimation but 
does not work well for 
dense %
reconstruction, and %
thus suffering from outliers. %
In contrast, our method can achieve more robust and accurate 3D scene reconstruction, without requiring offline-acquired camera poses and camera intrinsic parameters.

For quantitative efficiency analysis, the runtime comparisons on 40 Intel Xeon Silver 4210 CPUs and an RTX 3090 Ti GPU are presented in Table~\ref{tab: runtime analysis}, which includes three representative scenes with 225, 48, and 1000 images, respectively. Our pipeline achieves state-of-the-art reconstruction while only taking about a quarter of an hour to optimize, even for 1000 images. Note that only the time of predicting depth and poses without RGB-D fusion are recorded.

\subsection{Ablation Study}
\label{sec: ablation study}

We carry out ablation experiments in terms of essential components of our algorithm to evaluate their effectiveness. %

\begin{table}[!t]
\renewcommand\arraystretch{1.1}
  \centering
  \caption{\textbf{Comparison of geometric consistency alignment mechanisms on the NYU  \cite{silberman2012indoor} dataset.} 
  ``w/ deform." means optimizing the flexible deformation map of RCVD  \cite{kopf2021rcvd} as a replacement for our alignment module. ``baseline" means directly optimizing the per-frame pixel-wise depth map without the affine-invariant depth priors.
  As a result, the employment of our geometric consistency alignment module %
  contributes 
  significantly  %
  to the optimization. }
  \vspace{-0.5 em}
  \resizebox{\linewidth}{!}{%
  \setlength{\tabcolsep}{0.8mm}{
  \begin{tabular}{@{}r|lr|lcr|lr@{}}
    \toprule
    \multirow{2}{*}{Method} & \multicolumn{2}{c|}{Depth} & \multicolumn{3}{c|}{Pose} & \multicolumn{2}{c}{Reconstruction} \\
    \cline{2-8}
     & AbsRel$\downarrow$ & $\delta_1$$\uparrow$ & ATE$\downarrow$ & RPE-T$\downarrow$ & RPE-R$\downarrow$ & C-$l_1$$\downarrow$ & F-score$\uparrow$ \\
     
     \hline
     
     Ours & 0.092 & 0.923 & 0.096 & 0.144 & 0.053 & 0.099 & 0.622 \\
     
     \hline
      
     w/ deform.  \cite{kopf2021rcvd} & 0.153 & 0.771 & 0.145 & 0.290 & 0.082 & 0.160 & 0.470 \\
     
    \hline
    
    baseline & 0.350  & 0.466  & 0.710  & 1.097  & 0.215  & 0.285  & 0.303  \\
            
    \bottomrule
  \end{tabular}}
  }
  % \vspace{-1 em}
  \label{tab: ablation_lwlr}
\end{table}

\begin{table}[!t]
\renewcommand\arraystretch{1.1}
  \centering
  \caption{\textbf{Comparison of optimization objectives on the ScanNet  \cite{dai2017scannet} dataset.} The photometric constraint supervises the color consistency, the geometric constraint ensures the multi-view geometry consistency, and the regularization constraint stabilizes the convergence of optimization. As a result, the three components of our optimization objectives are all important.}
  \vspace{-0.5 em}
  \resizebox{\linewidth}{!}{%
  \setlength{\tabcolsep}{0.8mm}{
  \begin{tabular}{@{}r|lr|lcr|lr@{}}
    \toprule
    \multirow{2}{*}{Method} & \multicolumn{2}{c|}{Depth} & \multicolumn{3}{c|}{Pose} & \multicolumn{2}{c}{Reconstruction} \\
    \cline{2-8}
     & AbsRel$\downarrow$ & $\delta_1$$\uparrow$ & ATE$\downarrow$ & RPE-T$\downarrow$ & RPE-R$\downarrow$ & C-$l_1$$\downarrow$ & F-score$\uparrow$ \\
     
    \hline
    Ours & 0.127 & 0.858 & 0.271 & 0.348 & 0.147 & 0.170 & 0.410 \\
      
    \hline
      
    w/o photo. & 0.219 & 0.636 & 1.722 & 2.331 & 1.039 & 0.775 & 0.169 \\
    	
    w/o geo. & 0.168 & 0.750 & 0.316 & 0.404 & 0.179 & 0.211 & 0.323 \\
    
    w/o regu. & 0.226 & 0.603 & 0.245 & 0.338 & 0.189 & 0.220 & 0.254 \\

    \bottomrule
  \end{tabular}}
  }
  % \vspace{-2 em}
  % \vspace{-1 em}
  \label{tab: ablation_loss}
\end{table}

\noindent\textbf{Geometric Consistency Alignment Module.}
Table~\ref{tab: ablation_lwlr} shows the performance of different geometric consistency alignment modules. Previous consistent video depth methods RCVD and GCVD employ flexible deformation maps (``w/ deform.") and aim for more accurate scale-consistent depth. However, they both ignore the shift issues in affine-invariant depth, which is pretty significant in the 3D reconstruction task.
By contrast, our method (``Ours") proposes a global and local scale shift alignment module for depth rectification. We also compare with a baseline method (``baseline"), which directly optimizes all frames' pixel-wise depths without the affine-invariant depth priors. All depths are initialized to 1. Our alignment module works much better than previous methods. %

\noindent\textbf{Optimization Objectives.}
To evaluate the effectiveness of each constraint employed in our method, we propose to remove them one by one, and the results are shown in Table~\ref{tab: ablation_loss}. %
We can see that without the photometric constraint or geometric constraint, the performance will degrade a lot. When removing the regularization term, the accuracy of the pose, depth, and reconstruction will also decrease. 

\noindent\textbf{Effectiveness of Optimization Stages.}
Our optimization algorithm is composed of local keyframes optimization (``Local'') and global keyframes optimization (``Global''). The local stage endures the local consistency between the nearest $k$ keyframes, and can reconstruct roughly accurate 3D scene shapes. The global stage selects long-range keyframes to supervise global consistency. As shown in Table~\ref{tab: ablation_stages}, the reconstruction performance improves gradually.

\section{Conclusion}
In this paper, we have presented an effective pipeline to realize 3D scene reconstruction by leveraging the robustness of affine-invariant depth estimation, freezing the depth model, and jointly optimizing dozens of depth and camera parameters for each frame. Due to the sparsity of parameters, our pipeline can transfer the robustness of seemingly weak depth geometry prior to diverse scenes.
Extensive experiments show that our pipeline can achieve robust dense 3D reconstruction on challenging unseen scenes.

\begin{table}[!t]
\renewcommand\arraystretch{1.1}
  \centering
  \caption{\textbf{Effectiveness of our two-stage keyframe sampling and optimization strategy.} The ``Local'' stage ensures consistency between the nearest $k$ keyframes. The ``Global'' stage samples keyframes globally with different sampling probabilities according to the optimized relative poses. As a result, with the coarse-to-fine optimization strategy, our algorithm achieves better performance. The experiment is performed on NYU  \cite{silberman2012indoor} dataset.}
  % \vspace{-1 em}
  \resizebox{\linewidth}{!}{%
  \setlength{\tabcolsep}{0.8mm}{
  \begin{tabular}{@{}r|lr|lcr|lr@{}}
    \toprule
    \multirow{2}{*}{Stage} & \multicolumn{2}{c|}{Depth} & \multicolumn{3}{c|}{Pose} & \multicolumn{2}{c}{Reconstruction} \\
    \cline{2-8}
     & AbsRel$\downarrow$ & $\delta_1\uparrow$ & ATE$\downarrow$ & RPE-T$\downarrow$ & RPE-R$\downarrow$ & C-$l_1$ $\downarrow$ & F-score$\uparrow$ \\
          
     \hline
           
      Local & 0.111  & 0.884  & 0.111  & 0.179  & 0.068  & 0.091  & 0.593  \\
      Global & 0.092  & 0.923  & 0.096  & 0.144  & 0.053  & 0.099  & 0.622  \\

    \bottomrule
  \end{tabular}}
  }
  \label{tab: ablation_stages}
  % \vspace{-1 em}
\end{table}

\section*{Acknowledgements}
This work was supported by the National Key R\&D Program of China (No.\  2022ZD0118700), the JKW Research Funds under Grant 20-163-14-LZ-001-004-01, and the Anhui Provincial Natural Science Foundation under Grant 2108085UD12. We acknowledge the support of the GPU cluster built by MCC Lab of Information Science and Technology Institution, USTC.

{\small
\bibliographystyle{ieee_fullname}
\bibliography{draft_nocomments}
}

\newpage

\appendix
\newpage

\section*{\Large\textbf{Supplementary Material}}

\setcounter{table}{0}
\setcounter{figure}{0}
\renewcommand{\thefigure}{S\arabic{figure}}
\renewcommand{\thetable}{S\arabic{table}}

\section{Details About the LWLR Module}

Given a globally aligned depth map $\boldsymbol{D}^g$ and sparse guided points $\boldsymbol{y}$, the LWLR module\cite{xu2022boosting} recovers a location-aware scale-shift map. Concretely, for each 2D coordinate $(u,v)$, the sampled globally aligned depth $\boldsymbol{d}$ can be fitted to the ground-truth depth $\boldsymbol{y}$ by minimizing the squared locally weighted $\ell_2$ distance, which is re-weighted by a diagonal weight matrix $\boldsymbol{W}_{u,v}$.
\def\R{{\cal R}}
\begin{equation}
\begin{gathered}
\label{eq: modified_lwlr}
    \mathop {\min }\limits_{\boldsymbol{\beta}_{u,v}} ~~  (\boldsymbol{y} - \boldsymbol{X} \boldsymbol{\beta}_{u,v})^\mathsf{T} \boldsymbol{W}_{u,v} (\boldsymbol{y} - \boldsymbol{X} \boldsymbol{\beta}_{u,v}) + \lambda \theta_{u,v}^2 \\
    \boldsymbol{W}_{u,v} = diag(w_{1}, w_{2}, ..., w_{m}), ~w_{i} = \frac{1}{\sqrt{2\pi}} \exp(-\frac{ {\rm dist}_{i}^2}{2 b^2}) \\
    \boldsymbol{\hat{\beta}}_{u,v}=(\boldsymbol{X}^\mathsf{T} \boldsymbol{W}_{u,v} \boldsymbol{X} + \boldsymbol{A})^{-1} \boldsymbol{X}^\mathsf{T} \boldsymbol{W}_{u,v} \boldsymbol{y} \\
    \boldsymbol{A} = \begin{bmatrix} \lambda & 0 \\ 0 & 0 \end{bmatrix}, ~\boldsymbol{D} = \boldsymbol{S} \odot \boldsymbol{D}^g + \boldsymbol{\Theta} \\
    \boldsymbol{X}=[\boldsymbol{d}^\mathsf{T}, \boldsymbol{1}] \ \in \R^{m\times 2},~~ \boldsymbol{\beta}_{u,v}=[s_{u,v}, \theta_{u,v}]^\mathsf{T} \in \R^{2\times 1} \\
    \boldsymbol{D}, \boldsymbol{S}, \boldsymbol{D}^g, \boldsymbol{\Theta} \in \R^{H\times W}, 
    ~~ \boldsymbol{d}, \boldsymbol{y} \in \R^{m\times 1}  \\
\end{gathered}
\end{equation}
where $\boldsymbol{y}$ is the sampled sparse ground-truth metric depth, $\boldsymbol{d}$ is the sampled globally aligned depth, whose 2D coordinates are the same with those of sparse guided points $\boldsymbol{y}$. $\boldsymbol{X}$ is the homogeneous representation of $\boldsymbol{d}$, $m$ stands for the number of sampled points. $b$ is the bandwidth of Gaussian kernel, and $ {\rm dist}_{i}$ is the Euclidean distance between the the coordinate $(u_i,v_i)$ of $i$-th guided point and target point $(u,v)$. $\lambda$ is a $l_2$ regularization hyperparameter used for restricting the solution to be simple. By iterating the target point $(u,v)$ over the whole image, the scale map $\boldsymbol{S}$ and shift map $\boldsymbol{\Theta}$ can be generated composed of the scale values $s_{u,v}$ and shift values $\theta_{u,v}$ of each location $(u,v)$. Finally, the locally recovered metric depth $\boldsymbol{\hat{D}}$ equals to the shift map $\boldsymbol{\Theta}$ plus the Hadamard product~($\odot$, known as element-wise product) of the affine-invariant depth $\boldsymbol{D}$ and the scale map $\boldsymbol{S}$. The operation above can be summarized as below.
\begin{equation}
\begin{gathered}
\label{eq: f_lwlr}
    \boldsymbol{S}, \boldsymbol{\Theta} = f_{\text{LWLR}}(\boldsymbol{D}^g, \boldsymbol{y}) \\
    \boldsymbol{D} = \boldsymbol{S} \odot \boldsymbol{D}^g + \boldsymbol{\Theta}
\end{gathered}
\end{equation}

Rather than relying on sparse ground-truth metric depth $\boldsymbol{y}$, we replace it with $\{\omega_{i, t}\cdot d^{g}_i ( {\mathbf{p}_{t}} )\}^{M}_{t=1}$, which is related to parameters $\{\omega_{i, t}\}_{t=1}^{M}$ and sampled sparse global depth $\{d^{g}_i ( {\mathbf{p}_{t}} )\}^{M}_{t=1}$. By ensuring multi-frame consistency, we can retrieve scale-consistent depth maps.
\begin{equation}
\label{eq: depth alignment module supp}
\begin{gathered}
    \mathbf{A}_i, \mathbf{B}_i= f_{\text{LWLR}}(\mathbf{D}^g_i, \{\omega_{i, t} \cdot  d^{g}_i ( {\mathbf{p}_{t}} ) \}_{t=1}^M )  \\
    \mathbf{D}_i = \mathbf{A}_i \odot \mathbf{D}^g_i + \mathbf{B}_i
\end{gathered}
\end{equation}

\section{Efficiency of Photometric Constraint}

Furthermore, we also explore the efficiency of the photometric constraint on the NYU\cite{silberman2012indoor} dataset. We supervise the consistency of coordinates warped by the optical flow and the optimized parameters, as a replacement for photometric constraint. As shown in Table~\ref{tab: flow_loss}, although comparable performance can the flow-guided constraint achieve, it relies on predicting dense optical flow between every two frames with a robust model RAFT\cite{teed2020raft}, which can be time-consuming due to the recurrent refinement model and the $O(N^2)$ time complexity. In contrast, the photometric constraint does not require offline-computed optical flow and can be more efficient, especially on long-form videos.

\section{Keyframe Sampling Strategy}
We sample the keyframes with the optical flow computed with RAFT, and compare it with ours on the NYU dataset. The flow-guided keyframes will be selected if the valid regions are larger than 30 percent after checking the forward-backward consistency. As shown in Table~\ref{tab: ablation_raft_keyframe}, our algorithm is more efficient and even outperforms the flow-guided keyframe sampling strategy due to the pose-based long-range keyframe sampling.

\begin{table}[!t]
\renewcommand\arraystretch{1.1}
% \vspace{-1 em}
%  \newcommand{\tabincell}[2]{\begin{tabular}{@{}#1@{}}#2\end{tabular}}
  \centering
  \caption{\textbf{Efficiency of the photometric constraint.} We simply replace the photometric constraint with spatial constraint (``w/ flow''), and supervises the coordinates consistency warped by optical flow\cite{teed2020raft} and optimized parameters. Although comparable performance can it achieve, predicting dense optical flows with a deep network between every two frames is computationally expensive and time-consuming. Our algorithm can be more efficient while remains comparable performance.}
  % \vspace{-1 em}
  \resizebox{\linewidth}{!}{%
  \setlength{\tabcolsep}{0.8mm}{
  \begin{tabular}{@{}r|c|lr|lcr|lr@{}}
    \toprule
    \multirow{2}{*}{Method} & Time & \multicolumn{2}{c|}{Depth} & \multicolumn{3}{c|}{Pose} & \multicolumn{2}{c}{Reconstruction} \\
    \cline{3-9}
     & Complexity & AbsRel$\downarrow$ & $\delta_1\uparrow$ & ATE$\downarrow$ & RPE-T$\downarrow$ & RPE-R$\downarrow$ & C-$l_1$ $\downarrow$ & F-score$\uparrow$ \\
     
     \hline
      Ours & $O(1)$ & 0.092 & 0.923 & 0.096 & 0.144 & 0.053 & 0.099 & 0.622 \\
      
      \hline
      
      w/ flow & $O(N^2)$ & 0.102 & 0.907 & 0.095 & 0.155 & 0.052 & 0.085 & 0.627 \\

    \bottomrule
  \end{tabular}}
  }
  \label{tab: flow_loss}
  % \vspace{-2 em}
\end{table}

\begin{table}[!t]
\renewcommand\arraystretch{1.1}
  \centering
  \caption{\textbf{
  Keyframe sampling strategy.} We 
  sample keyframes according to the valid regions of the estimated optical flow. As a result, our algorithm not only spends less time but also achieves better performance.}
  % \vspace{-1 em}
  %
  \resizebox{\linewidth}{!}{%
  \setlength{\tabcolsep}{0.8mm}{
  \begin{tabular}{@{}r|c|lr|lcr|lr@{}}
    \toprule
    \multirow{2}{*}{Method} & Time & \multicolumn{2}{c|}{Depth} & \multicolumn{3}{c|}{Pose} & \multicolumn{2}{c}{Reconstruction} \\
    \cline{3-9}
     & Complexity & AbsRel$\downarrow$ & $\delta_1\uparrow$ & ATE$\downarrow$ & RPE-T$\downarrow$ & RPE-R$\downarrow$ & C-$l_1$ $\downarrow$ & F-score$\uparrow$ \\
     
     \hline
     
     Ours & $O(1)$ & 0.092 & 0.923 & 0.096 & 0.144 & 0.053 & 0.099 & 0.622 \\
     
     \hline
      
     w/ flow keyframe & $O(N^2)$ & 0.103  & 0.897  & 0.174  & 0.262  & 0.113  & 0.155  & 0.565  \\

    \bottomrule
  \end{tabular}}
  }
  % \vspace{-1 em}
  \label{tab: ablation_raft_keyframe}
\end{table}

\section{Upper Bound Analysis} 
Our optimization can work better if accurate GT poses and intrinsics are given. As shown in Table~\ref{tab: ablation_gt_intrinsic_pose}, the performance on the NYU dataset remains nearly the same with known GT intrinsic. With known GT poses, the quality of depth, pose, and reconstruction can be improved compared to video-only optimization. With both GT poses and intrinsic, the performance can achieve slightly better results.

\begin{table}[!t]
\renewcommand\arraystretch{1.1}
  \centering
  \caption{\textbf{Upper bound analysis.} With known GT poses, our algorithm can achieve better performance. The GT intrinsic 
  alone 
  does not improve the reconstruction performance
  , but can achieve slight performance improvement together with GT poses.}
  % \vspace{-1 em}
  %
  \resizebox{\linewidth}{!}{%
  \setlength{\tabcolsep}{0.8mm}{
  \begin{tabular}{@{}cc|cc|ccc|cc@{}}
    \toprule
    \multirow{2}{*}{GT intrinsic} & \multirow{2}{*}{GT poses} & \multicolumn{2}{c|}{Depth} & \multicolumn{3}{c|}{Pose} & \multicolumn{2}{c}{Reconstruction} \\
    \cline{3-9}
    & & AbsRel$\downarrow$ & $\delta_1\uparrow$ & ATE$\downarrow$ & RPE-T$\downarrow$ & RPE-R$\downarrow$ & C-$l_1$ $\downarrow$ & F-score$\uparrow$ \\
     
     \hline
     
       &  & 0.092 & 0.923 & 0.096 & 0.144 & 0.053 & 0.099 & 0.622 \\
       \checkmark &  & 0.092 & 0.920 & 0.095 & 0.147 & 0.050 & 0.103 & 0.622  \\
       & \checkmark & 0.085 & 0.933 & - & - & - & 0.070 & 0.662 \\
      \checkmark & \checkmark & 0.081 & 0.938 & - & - & - & 0.064 & 0.674 \\

    \bottomrule
  \end{tabular}}
  }
  % \vspace{-2 em}
  \label{tab: ablation_gt_intrinsic_pose}
\end{table}

\section{Analysis of Optimization Objectives}
Our optimization objectives are composed of photometric constraint, geometric constraint, and regularization constraint.
The photometric constraint $L_{pc}$ ensures the color consistency between the reference frame and the warped source frame. If we directly optimize the per-frame pixel-wise depth map with the photometric constraint, the supervision signal can be too weak to achieve satisfactory performance, especially on some low-texture regions. Here weak means supervising the color consistency instead of precise coordinate correspondences. However, the weak supervision becomes an advantage when it is employed with the geometric constraint and the robust affine-invariant depth prior. The affine-invariant depth maps can offer reliable inherent geometry information and narrow the solution space together with the geometric constraint. Concretely, the photometric constraint offers accurate guidance on the rich texture regions. For some low-texture regions, the photometric constraint will be small, and the optimization is mainly guided by the supervision of geometric consistency, which is also reliable due to the geometric accuracy of affine-invariant depth. 

For the geometric constraint $L_{gc}$, it can ensure the multi-view geometric consistency and will not bring any incorrect supervision, but the weight should not be too large to prevent from encouraging the whole depth map to be infinitely large. The weakly normalization supervision on the sparse guided points $L_{regu}$ is utilized to avoid extreme point cloud distortion and stabilize the optimization. 

\section{Discussion of Imperfect Cases}

Unlike other scenes, the outdoor KITTI sequences involve mostly straight-line movement with small differences between frames. During optimization, the photometric and geometric losses remain small even without accurate depths and poses. Although inexact, the depths and poses are consistent to enable acceptable reconstruction.

Despite the imperfect estimation of camera poses, we can still yield satisfactory reconstruction results, because it depends on the accuracy and the consistency of depth maps and poses. The advantage of our optimization pipeline lies in enabling the practical use of affine-invariant depth, and ensuring the aforementioned consistency. Also, the robustness of affine-invariant depth is transferred to pose estimation, leading to fewer failure cases such as `scene0707\_00'. Our pipeline also allows users to input offline-obtained poses, such as SfM poses, and jointly optimize for further improvement.

After optimization, the reconstructed point cloud and trajectory are consistent but still up to an unknown scale w.r.t. the real world. It is an intrinsic limitation of purely monocular reconstruction methods. The unknown scale can be recovered by providing the GT poses, or measuring the length of an object and aligning the reconstructed object’s size.

\begin{table*}[t]
\newcommand{\tabincell}[2]{\begin{tabular}{@{}#1@{}}#2\end{tabular}}
  \centering
  \caption{\textbf{Evaluation sequences of five zero-shot testing datasets.} Note that we evaluate the first sequences of 7-Scenes\cite{shotton2013scene}.}
  % \vspace{-1 em}
 \resizebox{\linewidth}{!}{%
\small 
  \begin{tabular}{@{}r|c@{}}
    \toprule

    Datasets & Scenes   \\
    
    \hline
    \multirow{2}{*}{NYU\cite{silberman2012indoor}}  & basement\_0001a, bedroom\_0015, bedroom\_0036, bedroom\_0059, classroom\_0004, computer\_lab\_0002, dining\_room\_0004, \\
    & dining\_room\_0033, home\_office\_0004, kitchen\_0008, kitchen\_0059, living\_room\_0058, office\_0006, office\_0024, playroom\_0002  \\

    \hline
    \multirow{2}{*}{ScanNet\cite{dai2017scannet}}       & scene0707\_00, scene0708\_00, scene0709\_00, scene0710\_00, scene0711\_00, scene0712\_00, scene0713\_00, \\ 
    & scene0714\_00, scene0715\_00, scene0716\_00, scene0717\_00, scene0718\_00, scene0719\_00, scene0720\_00  \\

    \hline
    7-Scenes\cite{shotton2013scene}    &  chess, fire, heads, office, pumpkin, redkitchen, stairs \\

    \hline
    TUM\cite{sturm2012benchmark}       & 360, desk, desk2, floor, plant, room, rpy, teddy, xyz  \\

    \hline
    KITTI\cite{geiger2013vision}  & 2011\_09\_26\_0001\_sync, 2011\_09\_26\_0009\_sync, 2011\_09\_26\_0091\_sync, 2011\_09\_28\_0001\_sync, 2011\_09\_29\_0004\_sync, 2011\_09\_29\_0071\_sync  \\
     
    \bottomrule
  \end{tabular}
 }
  \label{tab: eval_seq_list}
  % \vspace{-1 em}
\end{table*}

\section{Evaluation Details}

For 3D scene reconstruction, we %
evaluate the Chamfer $l_1$ distance (C-$l_1$) and F-score with a threshold of 5cm on the point cloud. %
Because of the unknown scale of estimated point clouds, we propose first to align the scale of depth maps and poses with ground truth through a global sharing scale factor, which is the ratio of the median depth value of all frames between optimized depth maps and ground-truth depth maps. Then, we match the estimated poses with ground truth through a $4\times4$ transformation matrix. The matrix is computed by employing Open3D's iterative closest point (ICP)\cite{besl1992method} algorithm between the optimized and the ground-truth point clouds. To reduce the negative effect of outliers for ICP matching, we remove some noisy points whose AbsRel errors are greater than $20\%$.

When evaluating depth, absolute relative error (AbsRel$=\frac{\left| d_{pred} - d_{gt} \right|}{d_{gt}}$) and the percentage of accurate depth pixels with $\delta_1=\max \bigl( \frac{d_{pred}}{d_{gt}}, \frac{d_{gt}}{d_{pred}}\bigr)<1.25$ are employed. %
To compare the consistency of depths along the video, we align all frames' depths with a global sharing factor. Similar to the 3D reconstruction evaluation, the scale factor is obtained by the ratio of the median depth value of all frames' depths between predictions and ground truths. %
For pose estimation, we follow \cite{sturm2012benchmark} to evaluate the 
absolute trajectory error (ATE), %
relative pose error of rotation (RPE-R) and translation (RPE-T). %
Before evaluation, the predicted poses are globally aligned with the ground truth.

For camera intrinsic, we evaluate the accuracy with the ``FOV AbsRel'', which is defined as the absolute relative error of the field of view (FOV AbsRel$=\frac{\left| \text{FOV}_{pred} - \text{FOV}_{gt} \right|}{\text{FOV}_{gt}}$).

\section{Optimization Details}

\noindent\textbf{Frames downsampling.} %
We propose a two-stage frame downsampling strategy to reduce the optimization time complexity. First, all frames $\{\mathbf{I}_i\}^{N}_{i=1}$ are fed to the LeReS-ResNeXt-101 network and get the backbone's last layer feature (the last layer of $1/32$ stage features) as their embeddings $\{\mathbf{e}_i\}_{i=1}^{S} \in \mathbb{R}^{N\times C\times{H}/{32}\times{W}/{32}}$. The first frame $\mathbf{I}_0$ is selected. We compute the similarity between $\mathbf{e}_0$ and its next neighboring $20$ frames $\{\mathbf{e}_i\}_{i=1}^{20}$, and each similarity between two frames is computed by constructing a 4D ${H}/{32}\times {W}/{32}\times {H}/{32}\times {W}/{32}$ cosine similarity volume of all pairs of two feature maps (similar to RAFT \cite{teed2020raft}) and take the maximum value as the image similarity. 

If a frame's similarity is just lower than a threshold value $\sigma$, then it is selected. We iteratively perform this process to sample several frames coarsely. In the second stage, we will evenly sample 3 frames between the first-stage adjacent samples. All sampled frames $\{\mathbf{I}_i\}^{P}_{i=1}$ are employed for next keyframes sampling and optimization. We set $\sigma$ to 0.85.

\noindent\textbf{Hyperparameters.} %
In the local stage, we use PyTorch's AdamW to optimize all learnable parameters. We iterate $2000$ steps in total. In each step, we random sample $50$ reference frames, and their paired keyframes are sampled based on $p_l$ for optimization. %
$\lambda_{pc}$, $\lambda_{gc}$, $\lambda_{regu}$ are set to $2$, $0.5$, and $0.01$ for indoor scenes and $2$, $0.001$, and $0.01$ for outdoor scenes respectively.

In the global stage, We iterate $4000$ steps in total for the global stage. In each step, we randomly sample $50$ reference frames and the paired keyframes based on $p_g$. $\phi$ is set to $\pi/4$. %
For indoor scenes, the $\lambda_{pc}$, $\lambda_{gc}$, $\lambda_{regu}$ are set to $2, 1, 0.1$ in first 2000 iters and $2, 0.1, 0.1$ in the last 2000 iters. The $\lambda_{gc}$ is set to $0.001$ for ourdoor scenes. 

Besides, we also filter out the sky regions for outdoor scenes by predicting semantic segmentation with SegFormer-B3\cite{xie2021segformer} during optimization.

\section{Testing Datasets}

In our experiments, we perform the evaluation on five zero-shot datasets: NYU\cite{silberman2012indoor}, ScanNet\cite{dai2017scannet}, 7-Scenes\cite{shotton2013scene}, TUM\cite{sturm2012benchmark}, and KITTI\cite{geiger2013vision}. The evaluation sequences of five zero-shot datasets are shown in Table~\ref{tab: eval_seq_list}. Note that we only evaluate the first sequences of 7 scenes on 7-Scenes, and 15, 14, 9, and 6 scenes on NYU, ScanNet, TUM, and KITTI individually. 

\begin{figure*}[h]
\centering    %
\includegraphics[width=\linewidth]{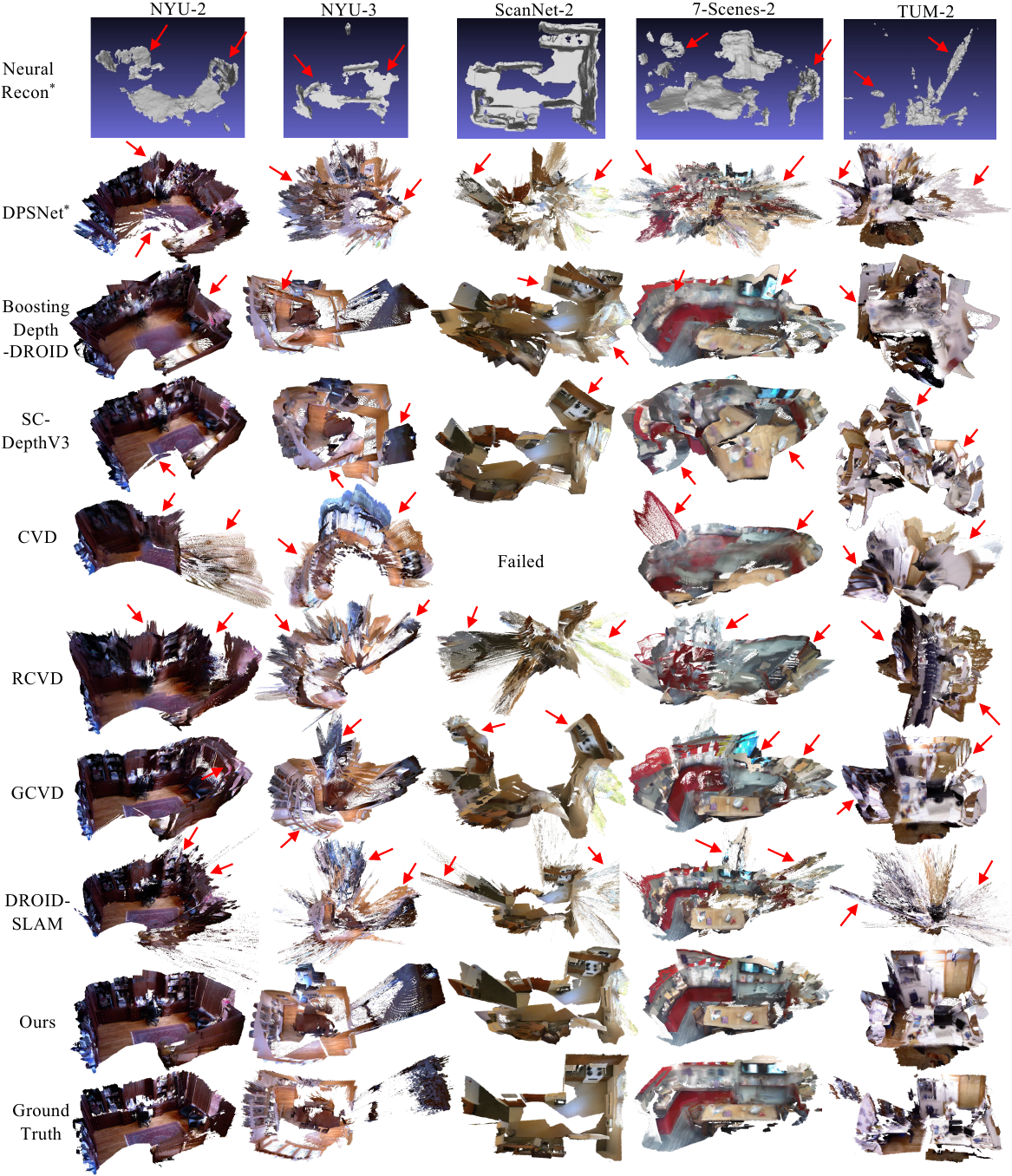}
% \vspace{-1 em}
\caption{\textbf{More qualitative comparisons of zero-shot 3D scene reconstruction.} Note that NeuralRecon is trained on ScanNet\cite{dai2017scannet} and can only output uncolored mesh, and $^*$ represents the employment of ground-truth camera poses during reconstruction.
} %
\label{fig: recon_supp}
% \vspace{-1 em}
\end{figure*}

\section{More Qualitative Comparisons}

More qualitative comparisons with seven representative algorithms are shown in Fig.~\ref{fig: recon_supp}. Our method can reconstruct accurate and robust 3D scene shapes on diverse scenes.

\end{document}